\documentclass[letterpaper]{article} 
\usepackage[]{aaai24}  
\usepackage{times}  
\usepackage{helvet}  
\usepackage{courier}  
\usepackage[hyphens]{url}  
\usepackage{graphicx} 
\usepackage{natbib}  
\usepackage{caption} 
\usepackage{algorithm}
\usepackage{algorithmic}

\usepackage{newfloat}
\usepackage{listings}

\usepackage{multicol, multirow, makecell, indentfirst, bm, booktabs, paralist, amssymb, amsmath}
\usepackage{colortbl}

\definecolor{mygray}{gray}{0.8}

\DeclareCaptionStyle{ruled}{labelfont=normalfont,labelsep=colon,strut=off} 
\floatstyle{ruled}
\newfloat{listing}{tb}{lst}{}
\floatname{listing}{Listing}
\title{Multi-Modal Domain Adaptation Across Video Scenes for Temporal Video Grounding}
\author{
    Haifeng Huang \hspace{1em} Yang Zhao \hspace{1em} Zehan Wang \hspace{1em} Yan Xia \hspace{1em} Zhou Zhao\\
}
\affiliations{
    Zhejiang University\\

    huanghaifeng@zju.edu.cn
}

\usepackage{bibentry}

\begin{document}

\maketitle

\begin{abstract}

    Temporal Video Grounding (TVG) aims to localize the temporal boundary of a specific segment in an untrimmed video based on a given language query. Since datasets in this domain are often gathered from limited video scenes, models tend to overfit to scene-specific factors, which leads to suboptimal performance when encountering new scenes in real-world applications. In a new scene, the fine-grained annotations are often insufficient due to the expensive labor cost, while the coarse-grained video-query pairs are easier to obtain. Thus, to address this issue and enhance model performance on new scenes, we explore the TVG task in an unsupervised domain adaptation (UDA) setting across scenes for the first time, where the video-query pairs in the source scene (domain) are labeled with temporal boundaries, while those in the target scene are not. Under the UDA setting, we introduce a novel Adversarial Multi-modal Domain Adaptation (AMDA) method to adaptively adjust the model's scene-related knowledge by incorporating insights from the target data. Specifically, we tackle the domain gap by utilizing domain discriminators, which help identify valuable scene-related features effective across both domains. Concurrently, we mitigate the semantic gap between different modalities by aligning video-query pairs with related semantics. Furthermore, we employ a mask-reconstruction approach to enhance the understanding of temporal semantics within a scene. Extensive experiments on Charades-STA, ActivityNet Captions, and YouCook2 demonstrate the effectiveness of our proposed method.

\end{abstract}

\section{Introduction\label{sec:intro}}
Temporal video grounding is a task that aims to locate the temporal boundary of a specific segment in an untrimmed video based on a given language query. This task has garnered increasing attention in recent years due to its potential applications, such as video search engines and automated video editing.

\begin{figure}[htb]
  \includegraphics[width=\linewidth]{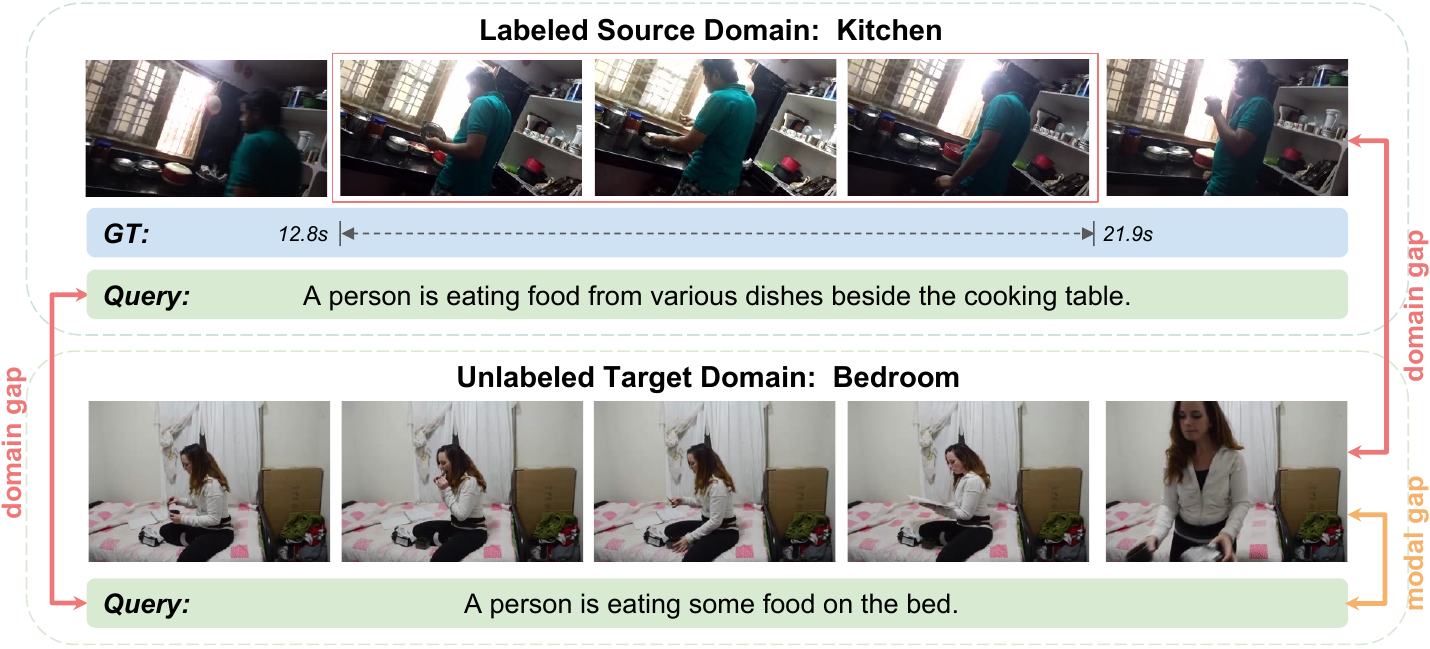}
    \caption{An example of unsupervised domain adaptation for TVG. The source domain \textit{Kitchen} is labeled with the ground truth (GT) of the temporal boundary, while the target domain \textit{Bedroom} is not.}
    \label{fig:intro}
  \end{figure}

Supervised methods require a large number of precise temporal boundary labels to train discriminative models, which is highly labor-intensive and time-consuming. Due to the expensive annotation cost, TVG datasets typically collect long untrimmed videos from only a single \cite{regneri2013grounding} or a few \cite{krishna2017dense, gao2017tall, zhou2018towards} scenes. These limited scenes impose strong scene-related factors on the learned knowledge of supervised models, leading to poor performance when applied directly to a new scene, as explicitly observed in our experiment.

Adapting the model to new scenes is a crucial aspect of this task in real-world applications. For instance, in the video surveillance field, the model must be capable of handling videos collected from various scenes. However, data from a new scene may be insufficient or lack fine-grained annotations, making it impractical to retrain a model on it. Additionally, it is currently infeasible to collect a sufficiently large dataset and train a scene-independent model.

However, compared to the expensive fine-grained annotations, the video-query pairs are usually easy to obtain from a new scene. In such scenarios, it is natural to approach this problem in a domain adaptation manner. Transferring knowledge from a labeled source domain to an unlabeled target domain is known as Unsupervised Domain Adaptation (UDA). In this work, we explore UDA across video scenes for the TVG task, where only the source domain/scene is labeled with temporal boundaries, while the target domain is not.

As an example, we consider two scenes (\textit{Kitchen} and \textit{Bedroom}) from the Charades-STA dataset, as shown in Figure~\ref{fig:intro}. For the action class "\textit{eat food}", a model trained on the \textit{Kitchen} scene would learn some \textit{Kitchen}-related knowledge, such as "the food is in a dish" or "the person is standing when eating". Consequently, when the model is applied to the \textit{Bedroom} scene, it may become confused about "a person eating on the bed" and experience a drop in performance. Under the UDA setting, the model can utilize data (video-query pairs) from the target domain to adaptively adjust the learned knowledge from the source domain. From another perspective, our goal is to reduce the domain gap of scene components across different domains and learn semantically-aligned representations across different modalities.

To address this problem, we propose the AMDA method to adaptively modulate the model's scene-related knowledge through a glimpse at the target data. Specifically, we tackle the aforementioned challenges in three ways: (i) we first reduce the domain gap by using adversarial domain discriminators on both single-modal and multi-modal features, which filters out useful scene factors that work well on both domains; (ii) we alleviate the semantic gap between different modalities by aligning semantically related video-query pairs; (iii) we employ a mask-reconstruction approach to further enhance the understanding of temporal semantics within a scene.

Our main contributions can be summarized as follows:

\begin{compactitem}
\item 
We present the first attempt to explore the unsupervised domain adaptation setting across video scenes for the temporal video grounding task, which poses a challenging multi-modal cross-domain problem.
\end{compactitem}
\begin{compactitem}
\item 
To enhance performance in new target scenes, we introduce a novel method called Adversarial-based Multi-modal Domain Adaptation (AMDA). This method adaptively adjusts the model's scene-related knowledge, effectively reducing both domain and modal gaps, all while maintaining temporal understandings.
\end{compactitem}
\begin{compactitem}
\item Extensive experiments conducted on Charades-STA, ActivityNet Captions, and YouCook2 demonstrate the effectiveness of our proposed model, which outperforms other traditional UDA methods.
\end{compactitem}

\section{Related Work}
\label{sec:related_work}

\paragraph{Temporal Video Grounding\label{sec:related_vtg}}
Temporal video grounding aims to locate the temporal boundaries of a specific segment in an untrimmed video based on a given language query. Most previous methods tackle the TVG task under the fully-supervised setting or weakly-supervised setting. The fully-supervised methods \cite{ge2019mac, wang2020temporally, zhang2020learning, zhang2019cross, xu2019multilevel, chen2020rethinking, mun2020local, nan2021interventional, rodriguez2020proposal, zeng2020dense, zhao2021cascaded, zhang2020span, xu2022hisa} require both video-query pairs and labeled temporal boundaries, which have been deeply explored and have achieved attractive performance. However, the high dependence on precise temporal boundary labels makes the labeling work extremely time-consuming and labor-intensive. Therefore, researchers begin to solve this problem under the weakly-supervised setting \cite{mithun2019weakly, chen2020look, lin2020weakly, teng2021regularized, zheng2022weakly}, where the models only need video-query pairs without the annotations of boundary time. Besides, some researchers~\cite{wang2023scene, fang2022multi} start to explore domain-invariant and transferable methods for this task.

\paragraph{Unsupervised Domain Adaptation}
Unsupervised domain adaptation (UDA) aims to transfer knowledge from the labeled source domain to the unlabeled target domain. The source and target domains have similar but not identical distributions. Recent works can be roughly categorized into two groups: \textit{moment matching-based} methods and \textit{adversarial-based} methods. The \textit{moment matching-based} methods try to alleviate the domain gap by aligning the statistics of source and target distributions, such as MMD \cite{tzeng2014deep, long2015learning, pan2010domain}, CORAL \cite{sun2017correlation}, and manifold metrics \cite{luo2020unsupervised}.

The \textit{adversarial-based} methods use a domain discriminator to distinguish the source and target domain, such as DANN \cite{ganin2016domain} and ADDA \cite{tzeng2017adversarial}. After this, more and more works attempt to solve the UDA problem by minimizing the domain discrepancy at a domain-level \cite{yang2020heterogeneous, long2017deep, long2018conditional, ren2018generalized, zhu2020deep}. Following the mainstream methods of UDA, we also build an \textit{adversarial-based} architecture to reduce the domain gap between different domains.

In recent works, more and more attention has been given to deep UDA in video tasks \cite{chen2019temporal, jamal2018deep, munro2020multi, xu2022aligning, chen2022multi}. And some of them attempted multi-modal video-text tasks \cite{munro2021domain, chen2021mind}. However, when doing domain adaptation for videos, these multi-modal methods usually neglect the temporal axis of video representations. They work well in video-text retrieval tasks, but they cannot be directly applied to the TVG task to predict a precise temporal boundary.

\begin{figure*}[htb]
    \includegraphics[width=\linewidth]{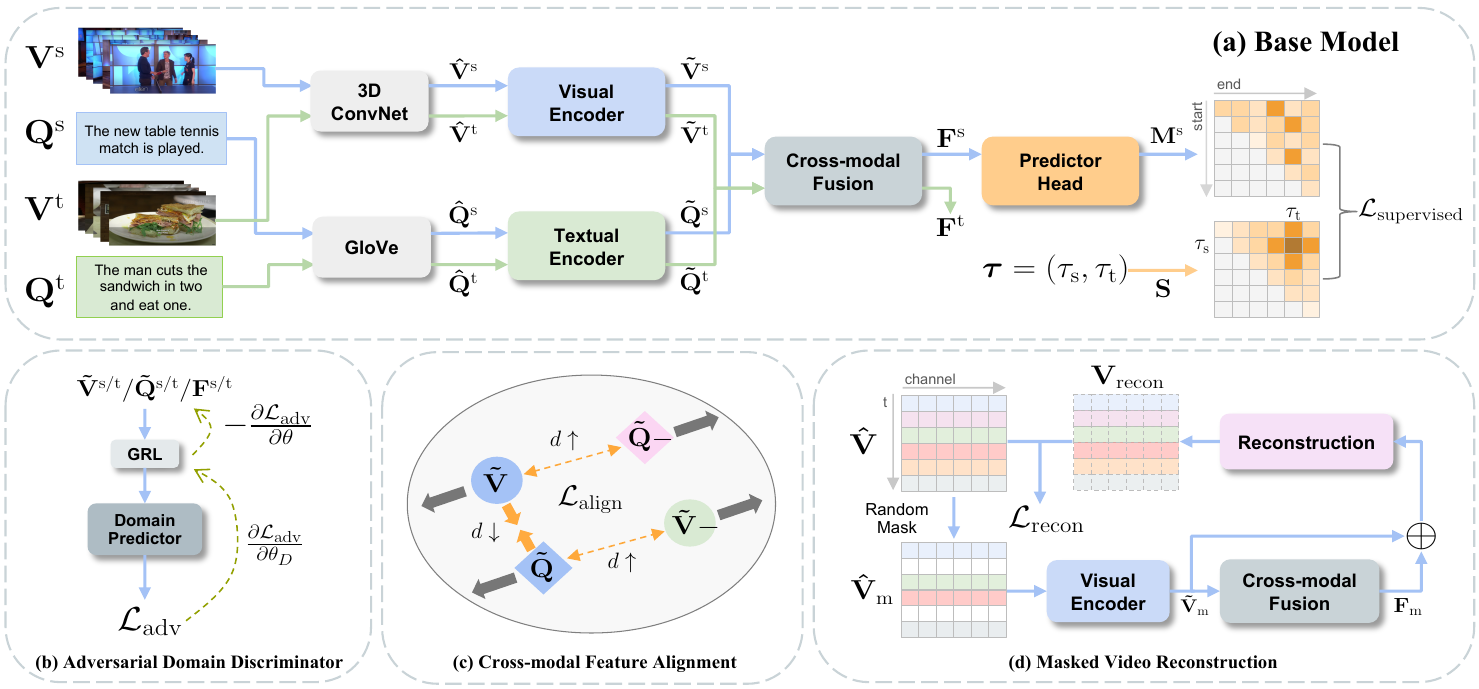}
    \caption{The overall architecture of the AMDA model. (a) The base model encodes and fuses the visual and textual features, and the predictor head is for supervised training on the source domain. (b) We reduce the domain gap with a domain discriminator. (c) We align the video-query pairs in the semantic space. (d) We learn the temporal semantics in a mask-reconstruction manner.}
  \label{fig:model}
\end{figure*}

\section{Proposed Method}
In this section, we first give the problem formulation of this task. Then we introduce the Base Model, which contains the feature encoders, cross-modal fusion, and the predictor head, as shown in Figure~\ref{fig:model}(a). The predictor head is used for supervised training on the source domain. 

Based on the base model, we propose the Adversarial Multi-modal Domain Adaptation (AMDA) model, which aims to transfer knowledge from the source domain to the target domain. It comprises three modules: adversarial domain discriminator, cross-modal feature alignment, and masked video reconstruction, as shown in Figure~\ref{fig:model}(bcd).

\subsection{Problem Formulation\label{sec:form}}
Given an untrimmed video $\mathbf{V}=\{\mathbf{v}_i\}_{i=1}^{n}$ and a corresponding language query $\mathbf{Q}=\{\mathbf{q}_i\}_{i=1}^{m}$, TVG aims to localize a temporal boundary $\boldsymbol{\tau}=(\tau_\mathrm{s},\tau_\mathrm{e})$ starting at timestamp $\tau_\mathrm{s}$ and ending at timestamp $\tau_\mathrm{t}$ in video $\mathbf{V}$, which matches the language query $\mathbf{Q}$ semantically. $n$ and $m$ are the frame number of video and the length of query, respectively.

In the UDA setting, we are given a set of untrimmed videos $\mathbb{V}^\mathrm{s}$/$\mathbb{V}^\mathrm{t}$ and language queries $\mathbb{Q}^\mathrm{s}$/$\mathbb{Q}^\mathrm{t}$ from the source/target domain. The video-query pairs are labeled with the temporal boundaries $\mathbb{T}^\mathrm{s}$ in the source domain but not in the target one. Under such circumstances, the main objective is to derive an effective boundary predictor on the target domain by fully exploiting labeled data in the source domain and unlabeled data in the target domain. Details will be illustrated in Section~\ref{sec:adma}.

\subsection{Base Model\label{sec:base}}

\paragraph{\textbf{Feature Encoder}}
Given an input untrimmed video $\mathbf{V}$ and a corresponding language query $\mathbf{Q}$, we first utilize the pre-trained 3D ConvNets \cite{tran2015learning} to extract visual features $\mathbf{\hat{V}}$ and the 300d GloVe embedding \cite{pennington2014glove} to extract textual features $\mathbf{\hat{Q}}$ to keep consistent with other methods. After projecting them into the same hidden dimension, we employ multi-head self-attention layers to encode them. We denote the encoded visual and textual features as $\mathbf{\tilde{V}}=\{\mathbf{\tilde{v}}_i\}_{i=1}^{n}\in\mathbb{R}^{n\times d}$ and $\mathbf{\tilde{Q}}=\{\mathbf{\tilde{q}}_i\}_{i=1}^{m}\in\mathbb{R}^{m\times d}$ , respectively. $d$ is the hidden dimension.

\paragraph{\textbf{Cross-modal Fusion}}
We utilize a Context-Query Attention \cite{yu2018qanet} module to fuse the visual and textual features. Specifically, we first calculate the cross-modal similarity matrix $\mathbf{S}\in \mathbb{R}^{n\times m}$ via the cosine similarity function. And we normalize the similarity matrix along row axis and column axis, respectively, to get $\mathbf{S}_\mathrm{r}$ and $\mathbf{S}_\mathrm{c}$. Then the video-to-query attention $\mathbf{A}_\mathrm{v}\in \mathbb{R}^{n\times d}$ and the query-to-video attention $\mathbf{A}_\mathrm{q}\in \mathbb{R}^{m\times d}$ can be calculated as
\begin{equation}
\mathbf{A}_\mathrm{v}=\mathbf{S}_\mathrm{c}\cdot\mathbf{\tilde{Q}},\ \mathbf{A}_\mathrm{q}=\mathbf{S}_\mathrm{c}\cdot\mathbf{S}_\mathrm{r}^\mathrm{T}\cdot\mathbf{\tilde{V}}
\end{equation}
And the fused representation $\mathbf{F}\in\mathbb{R}^{n\times d}$ can be computed as
\begin{equation}
\mathbf{F}=([\mathbf{\tilde{V}};\,\mathbf{A}_\mathrm{v};\,\mathbf{\tilde{V}}\odot\mathbf{A}_\mathrm{v};\,\mathbf{\tilde{V}}\odot\mathbf{A}_\mathrm{q}])\mathbf{W}_\mathrm{f} + \mathbf{b}_\mathrm{f},
\end{equation}
where $\mathbf{W}_\mathrm{f}\in\mathbb{R}^{4d\times d}$ and $\mathbf{b}_\mathrm{f}\in\mathbb{R}^d$ are all learnable parameters of a linear layer.

After cross-modal fusion, many state-of-the-art supervised methods will construct a complex module to learn further the local-global contexts, while this is not the critical concern in our task. For simplicity, we only utilize a multi-head attention layer to obtain the final representation $\mathbf{\tilde{F}}={\{\mathbf{\tilde{f}}_i}\}_{i=1}^{n_v}\in\mathbb{R}^{n\times d}$, which is simple but also effective.

\paragraph{\textbf{Predictor Head}}

Following the former work \cite{liu2021context}, we adapt biaffine mechanism to this task for scoring all possible $(start, end)$ pairs. We first employ two separate feed-forward networks to generate hidden representations for start frames and end frames, respectively. Then we employ the biaffine operation over them to get the score map $\mathbf{M}=\{M_\mathbf{p}\}_{\mathbf{p}=(1,1)}^{(n,n)}\in\mathbb{R}^{n\times n}$.

For each segment $\mathbf{p}=(p_s,p_e)$, where $p_s$/$p_e$ is the start/end timestamp of segment $\mathbf{p}$, the calculating process is as follows:
\begin{equation}
\begin{aligned}
\label{eq:mp}
\mathbf{r}_\mathbf{p}^\mathrm{s}=\operatorname{FFN}^\mathrm{s}(\mathbf{\tilde{f}}_{p_s}),\ &\ \ \mathbf{r}_\mathbf{p}^\mathrm{e}=\operatorname{FFN}^\mathrm{e}(\mathbf{\tilde{f}}_{p_e}),\\
M_\mathbf{p}=\sigma(\mathbf{r}_\mathbf{p}^\mathrm{s}\mathbf{U}_\mathrm{m}(\mathbf{r}_\mathbf{p}^\mathrm{e})^\top+&(\mathbf{r}_\mathbf{p}^\mathrm{s}\oplus \mathbf{r}_\mathbf{p}^\mathrm{e})\mathbf{W}_\mathrm{m}+b_\mathrm{m}),
\end{aligned}
\end{equation}
where $\operatorname{FFN}^\mathrm{s}$ and $\operatorname{FFN}^\mathrm{e}$ are feed-forward networks. $\mathbf{U}_\mathrm{m}\in\mathbb{R}^{d\times d}$, $\mathbf{W}_\mathrm{m}\in\mathbb{R}^d$ and $b_\mathrm{m}\in\mathbb{R}$ are learnable parameters. $\oplus$ denotes element-wise addition. $\sigma$ is the sigmoid function. $\mathbf{r}_\mathbf{p}^\mathrm{s}$/$\mathbf{r}_\mathbf{p}^\mathrm{e}\in\mathbb{R}^d$ is the hidden representation of the start/end frame. Then $M_\mathbf{p}$ is the matching score of segment $\mathbf{p}$, which indicates the possibility of this segment matching the sentence query.

\subsection{Adversarial Multi-modal Domain Adaptation\label{sec:adma}}

When it comes to the UDA setting, for the sake of clarity in the following descriptions, we denote the input data from the source and target domains in a batch-like way, given by:
\begin{equation}\nonumber
\begin{aligned}
\mathbb{V}^\mathrm{s}&=\{\mathbf{V}_i^\mathrm{s}\}_{i=1}^{B},\ \mathbb{Q}^\mathrm{s}=\{\mathbf{Q}_i^\mathrm{s}\}_{i=1}^{B},\ \mathbb{T}^\mathrm{s}=\{\boldsymbol{\tau}_i^\mathrm{s}\}_{i=1}^{B} \\
\mathbb{V}^\mathrm{t}&=\{\mathbf{V}_i^\mathrm{t}\}_{i=1}^{B},\ \mathbb{Q}^\mathrm{t}=\{\mathbf{Q}_i^\mathrm{t}\}_{i=1}^{B},\ \mathbb{T}^\mathrm{t}=\varnothing
\end{aligned}
\end{equation}
where $\mathbf{V}_i^\mathrm{s}$/$\mathbf{V}_i^\mathrm{t}$ and $\mathbf{Q}_i^\mathrm{s}$/$\mathbf{Q}_i^\mathrm{t}$ are the $i$-th pair of the untrimmed video and the sentence query in the source/target domain, respectively. $\boldsymbol{\tau}_i^s$ is the labeled temporal boundary matched to the $i$-th sentence query in the source domain. $B$ is the batch size. 

After the feature encoder, the encoded visual and textual features can be denoted as:
\begin{equation}\nonumber
\begin{aligned}
\mathbb{\tilde{V}}^\mathrm{s}&=\{\mathbf{\tilde{V}}_i^\mathrm{s}\}_{i=1}^{B},\ \mathbb{\tilde{Q}}^\mathrm{s}=\{\mathbf{\tilde{Q}}_i^\mathrm{s}\}_{i=1}^{B},\\
\mathbb{\tilde{V}}^\mathrm{t}&=\{\mathbf{\tilde{V}}_i^\mathrm{t}\}_{i=1}^{B},\ \mathbb{\tilde{Q}}^\mathrm{t}=\{\mathbf{\tilde{Q}}_i^\mathrm{t}\}_{i=1}^{B}.
\end{aligned}
\end{equation}

Both domains will conduct cross-modal fusion, and the fused representation are $\mathbb{F}^\mathrm{s}=\{\mathbf{F}^\mathrm{s}\}_{i=1}^{B}$ and $\mathbb{F}^\mathrm{t}=\{\mathbf{F}^\mathrm{t}\}_{i=1}^{B}$.

\paragraph{\textbf{Adversarial Domain Discriminator}}
As discussed in Section~\ref{sec:intro}, different scene factors in the source and target domain will cause the domain shift. We propose using two domain discriminators, $\operatorname{D}_\mathrm{v}$ and $\operatorname{D}_\mathrm{q}$, respectively, for the visual and textual modalities to reduce the domain gap by confusing the domain discriminators so that they cannot correctly distinguish a visual/textual sample from the source domain or the target domain. 

For the encoded visual feature $\mathbf{\tilde{V}}^\mathrm{s}$/$\mathbf{\tilde{V}}^\mathrm{t}$ and textual feature $\mathbf{\tilde{Q}}^\mathrm{s}$/$\mathbf{\tilde{Q}}^\mathrm{t}$, the adversarial domain discriminators can be formulated as:
\begin{equation}\label{eq:dvdq}
\begin{aligned}
\operatorname{D}_\mathrm{v}(\mathbf{\tilde{V}}^\mathrm{k}) &= \sigma(\operatorname{MLP}_\mathrm{v}(\operatorname{GRL}(\mathbf{\tilde{V}}^\mathrm{k}))),\\
\operatorname{D}_\mathrm{q}(\mathbf{\tilde{Q}}^\mathrm{k}) &= \sigma(\operatorname{MLP}_\mathrm{q}(\operatorname{GRL}(\mathbf{\tilde{Q}}^\mathrm{k}))),
\end{aligned}
\end{equation}
where $\mathrm{k}\in\{\mathrm{s},\mathrm{t}\}$. $\operatorname{MLP}_\mathrm{v}$ and $\operatorname{MLP}_\mathrm{q}$ are two multi-layer perceptrons for visual and textual modality, respectively. $\operatorname{GRL}$ is a gradient reversal layer, where the gradient is reversed in back-propagation. $\sigma$ is the sigmoid function. 


However, the domain discriminators $\operatorname{D}_\mathrm{v}$/$\operatorname{D}_\mathrm{q}$ only help learn single-modal (visual/textual) domain-invariant features. To consider all modalities together, we utilize another $\operatorname{D}_\mathrm{f}$ to learn cross-modal domain-invariant representations. For a fused feature $\mathbf{F}^\mathrm{s}$/$\mathbf{F}^\mathrm{t}$,  the domain discriminator $\operatorname{D}_\mathrm{f}$ can be formulated as:
\begin{equation}\label{eq:df}
\operatorname{D}_\mathrm{f}(\mathbf{F}^\mathrm{k})=\sigma(\operatorname{MLP}_\mathrm{f}(\operatorname{GRL}(\mathbf{F}^\mathrm{k})),
\end{equation}
where $\mathrm{k}\in\{\mathrm{s},\mathrm{t}\}$. $\operatorname{MLP}_\mathrm{f}$ is a multi-layer perceptron and $\operatorname{GRL}$ is a gradient reversal layer after the cross-modal fusion module.

\paragraph{\textbf{Cross-modal Feature Alignment}}
In the unlabeled target domain, the visual and textual features are difficult to be semantically aligned without supervised training. Although the domain discriminator $\operatorname{D}_\mathrm{f}$ attempts to reduce the domain gap between fused features in the source and target domain, it is still an indirect way and cannot align the semantic information well between different modalities.

We propose the cross-modal feature alignment module based on the triplet loss. For a pair of visual-textual features $\mathbf{\overline{V}},\mathbf{\overline{Q}}\in\mathbb{R}^d$,  The general cross-modal alignment loss can be formulated as:
\begin{equation}\label{eq:align_loss}
\begin{aligned}
&\mathcal{L}_\mathrm{CA}(\mathbf{\overline{V}},\mathbf{\overline{Q}},\mathbb{\overline{V}}-,\mathbb{\overline{Q}}-)=\\
&\sum_{\mathbf{\overline{V}}-\in \mathbb{\overline{V}}-}\mathop{max}(0,\Delta-l(\mathbf{\overline{V}},\mathbf{\overline{Q}})+l(\mathbf{\overline{V}}-,\mathbf{\overline{Q}}))+\\
&\sum_{\mathbf{\overline{Q}}-\in \mathbb{\overline{Q}}-}\mathop{max}(0,\Delta-l(\mathbf{\overline{V}},\mathbf{\overline{Q}})+l(\mathbf{\overline{V}},\mathbf{\overline{Q}}-)),
\end{aligned}
\end{equation}
where $l(\cdot,\cdot)$ is a cosine similarity function. $\mathbf{\overline{V}},\mathbf{\overline{Q}}\in\mathbb{R}^d$ are a pair of positive features obtained by averaging along the temporal axis of encoded feature $\mathbf{\tilde{V}},\mathbf{\tilde{Q}}\in\mathbb{R}^{n\times d}$. $\mathbb{\overline{V}}-$ and $\mathbb{\overline{Q}}-$ are the sets of chosen negative samples for $\mathbf{\overline{V}}$ and $\mathbf{\overline{Q}}$, respectively. $\Delta$ is a margin. 

The goal of the $\mathcal{L}_\mathrm{CA}$ is to pull positive pairs closer and push negative pairs further in the semantic space. We apply this alignment loss to encourage the semantically related video-query pairs to be similar and reduce the similarity between unrelated negative pairs, as shown in Figure~\ref{fig:model}(c).

\paragraph{\textbf{Masked Video Reconstruction}} Unlike the video-text retrieval task that only needs to retrieve the whole video related to the text, the VTG task is to localize the precise temporal boundary of the video segment, which requires the model to focus on temporal semantics within a video.

However, in the above cross-modal feature alignment module, we align the visual and textual modalities through the mean value of encoded features, which lacks attention to temporal semantics and may cause inaccurate prediction in the target domain.

We propose the masked video reconstruction module to learn the temporal semantic relations and robust features. As shown in Figure~\ref{fig:model}(d), the key idea is to generate the masked feature $\mathbf{\hat{V}}_\mathrm{m}$ by randomly masking some frames of the video feature $\mathbf{\hat{V}}$ at a ratio of $\beta$. After obtaining the encoded feature $\mathbf{\tilde{V}}_\mathrm{m}$ and the fused feature $\mathbf{F}_\mathrm{m}$, the reconstructed feature $\mathbf{V}_\mathrm{recon}$ can be computed as:
\begin{equation}\label{eq:recon}
\mathbf{V}_\mathrm{recon}=\operatorname{Conv1D}(\operatorname{ReLU}(\operatorname{Conv1D}(\mathbf{\tilde{V}}_\mathrm{m}\oplus \mathbf{F}_\mathrm{m}))),
\end{equation}
where $\oplus$ denotes element-wise addition.

The masked video reconstruction not only urges the feature encoder to perceive and aggregate information from the visual context to predict the masked frames but also encourages the cross-modal fusion to get useful information from the sentence query to help complete the missing frames.

\begin{table*}[htb]
  \centering
  \caption{Performance comparison with UDA baselines on Charades-STA. (source domain: \textit{Kitchen})}
  \label{tab:charades_source_kitchen}

  \scalebox{0.9}{
  \begin{tabular}{c|cc|cc|cc|cc|cc}
    \toprule
    \multirow{2}{*}{Method} & \multicolumn{2}{c|}{$\rightarrow$ Bedroom} & \multicolumn{2}{c|}{$\rightarrow$ Living room} & \multicolumn{2}{c|}{$\rightarrow$ Bathroom} & \multicolumn{2}{c|}{$\rightarrow$ Entryway} & \multicolumn{2}{c}{Average}\\
    & IoU=0.5 & IoU=0.7
    & IoU=0.5 & IoU=0.7
    & IoU=0.5 & IoU=0.7
    & IoU=0.5 & IoU=0.7
    & IoU=0.5 & IoU=0.7
    \\
    \midrule

    \rowcolor{mygray}Supervised-target & 42.07 & 20.33 & 41.19 & 23.96 & 39.39 & 21.78 & 32.73 & 14.77 & 38.85 & 20.21 \\
    
    & & & & & & & & & & \\[-2ex]
    \hline 
    & & & & & & & & & & \\[-2ex]
    
    Source-only     & 28.82 & 13.31 & 36.85 & 20.96 & 35.58 & 17.01 & 21.15 & 11.22 & 30.60 & 15.63 \\
    MK-MMD          & 29.37 & 13.46 & 37.05 & 19.26 & 35.72 & 17.28 & 25.32 & 11.49 & 31.87 & 15.37 \\
    DeepCORAL          & 30.45 & 13.22 & 37.11 & 19.34 & 35.64 & 17.15 & 27.06 & 11.85 & 32.57 & 15.39 \\
    AFN       & 33.59 & 16.14 & 37.63 & 19.92 & 35.99 & 16.21 & 29.12 & 12.90 & 34.08 & 16.29 \\
    ACAN             & 33.62 & 15.60 & 38.35 & 19.73 & 36.59 & 17.30 & 28.31 & 12.94 & 34.22 & 16.39\\
    \hline 
    & & & & & & & & & &\\[-2ex]
    AMDA ($\mathcal{L}_\mathrm{adv}$ only) & 33.65 & 15.75 & 38.35 & 20.88 & 36.65 & 17.95 & 29.37 & 12.55 & 34.51 & 16.78 \\
    AMDA & \textbf{39.25} & \textbf{17.31} & \textbf{40.04} & \textbf{22.98} & \textbf{37.34} & \textbf{19.91} & \textbf{32.44} & \textbf{14.25} & \textbf{37.27} & \textbf{18.61} \\

    \bottomrule
  \end{tabular}}
\end{table*}

\subsection{Training and Inference\label{sec:train}}
We train our AMDA model end-to-end via a multi-task loss function, containing supervised loss $\mathcal{L}_\mathrm{sup}$ of the Base Model, adversarial loss $\mathcal{L}_\mathrm{adv}$ of domain discriminator, cross-modal alignment loss $\mathcal{L}_\mathrm{align}$ and reconstruction loss $\mathcal{L}_\mathrm{recon}$ of masked videos.

\paragraph{\textbf{Supervised Loss of Base Model}}
In the labeled source domain, we utilize the scaled Intersection over Union (IoU) values as the supervision signal. Given the ground truth $\boldsymbol{\tau}=(\tau_s,\tau_t)$, the scaled IoU score $s_\mathbf{p}$ of the segment $\mathbf{p}=(p_s,p_t)$ can be computed as
$s_\mathbf{p}=\operatorname{IoU}(\mathbf{p},\boldsymbol{\tau})/\operatorname{Max}(\mathbf{S})$,
where $\operatorname{Max}(\mathbf{S})$ is the maximum value of all IoU scores $\mathbf{S}=\{s_\mathbf{p}\}_{\mathbf{p}=(1,1)}^{(n,n)}\in\mathbb{R}^{n\times n}$ . Then the base model is trained by a binary cross-entropy loss:
\begin{equation}
\mathcal{L}_\mathrm{sup}=-\sum_{\mathbf{p}=(1,1)}^{(n,n)}s_\mathbf{p}\operatorname{log}(M_\mathbf{p})+(1-s_\mathbf{p})\operatorname{log}(1-M_\mathbf{p}),
\end{equation}
where $M_\mathbf{p}$ is the score of the segment $\mathbf{p}$ illustrated in Equation~\ref{eq:mp}.

\paragraph{\textbf{Adversarial Loss of Domain Discriminator}}
We define the domain labels in the source and target domain as $\operatorname{d}(\mathbf{\tilde{V}}^\mathrm{s}/\mathbf{\tilde{Q}}^\mathrm{s}/\mathbf{F}^\mathrm{s})=0$ and $\operatorname{d}(\mathbf{\tilde{V}}^\mathrm{t}/\mathbf{\tilde{Q}}^\mathrm{t}/\mathbf{F}^\mathrm{t})=1$, respectively.

For each kind of feature $\mathbf{E}\in\{\mathbf{\tilde{V}},\mathbf{\tilde{Q}},\mathbf{F}\}$, the adversarial loss of domain discriminator is:
\begin{equation}
    \mathcal{L}_\mathrm{adv}=\sum_{i=1}^{B}\operatorname{BCELoss}(\operatorname{D}_\mathrm{*}(\mathbf{E}^\mathrm{k}_i),\operatorname{d}(\mathbf{E}^\mathrm{k}_i))
\end{equation}
where $\mathrm{k}\in\{\mathrm{s},\mathrm{t}\}$, $\operatorname{BCELoss}$ is the binary cross entropy loss. $\operatorname{D}_\mathrm{*}$ is the corresponding domain discriminator defined in Equation~\ref{eq:dvdq} and \ref{eq:df}.

\paragraph{\textbf{Cross-modal Alignment Loss}}
Given the encoded features $\mathbb{\tilde{V}}^\mathrm{t}=\{\mathbf{\tilde{V}}^\mathrm{t}_i\}_{i=1}^B,\mathbb{\tilde{Q}}^\mathrm{t}=\{\mathbf{\tilde{Q}}^\mathrm{t}_i\}_{i=1}^B$ in the target domain, we first calculate the mean value $\mathbb{\overline{V}}^\mathrm{t}=\{\mathbf{\overline{V}}_i^\mathrm{t}\}_{i=1}^B,\mathbb{\overline{Q}}^\mathrm{t}=\{\mathbf{\overline{Q}}_i^\mathrm{t}\}_{i=1}^B\in\mathbb{R}^{B\times d}$ along the temporal axis. Then the cross-modal alignment loss can be calculated as:
\begin{equation}
\mathcal{L}_\mathrm{align}=\sum_{i=1}^B\mathcal{L}_\mathrm{CA}(\mathbf{\overline{V}}_i^\mathrm{t},\mathbf{\overline{Q}}_i^\mathrm{t},\mathbb{\overline{V}}_i^\mathrm{t}-,\mathbb{\overline{Q}}_i^\mathrm{t}-),
\end{equation}
where $\mathbb{\overline{V}}_i^\mathrm{t}-=\{\mathbf{\overline{V}}_j^\mathrm{t}|j\in[1,B]$ and $j\neq i\}$ and $\mathbb{\overline{Q}}_i^\mathrm{t}-=\{\mathbf{\overline{Q}}_j^\mathrm{t}|j\in[1,B]$ and $j\neq i\}$ are the negative samples for the $i$-th textual and visual feature, respectively. $\mathcal{L}_\mathrm{CA}$ is formulated in Equation~\ref{eq:align_loss}.

\paragraph{\textbf{Reconstruction Loss of Masked Videos}}
We reduce the distance between the reconstructed feature $\mathbf{V}_\mathrm{recon}$ and the originally extracted feature $\mathbf{\hat{V}}$ by:
\begin{equation}
\mathcal{L}_\mathrm{recon}=\operatorname{MSELoss}(\mathbf{V}_\mathrm{recon},\mathbf{\hat{V}}),
\end{equation}
where $\operatorname{MSELoss}$ is the mean squared error loss.

To be consistent with the adversarial domain discriminator module, the encoded feature $\mathbf{\tilde{V}}_\mathrm{m}$ and fused feature $\mathbf{F}_\mathrm{m}$ of masked videos are also sent into the domain discriminators $\operatorname{D}_\mathrm{v}$ and $\operatorname{D}_\mathrm{f}$, respectively, to learn the domain-invariant representations.

\paragraph{\textbf{Total Loss}}
Finally, the overall loss function in the training process can be formulated as:
\begin{equation}
\mathcal{L}_\mathrm{total}=\mathcal{L}_\mathrm{sup}+\lambda_1\mathcal{L}_\mathrm{adv}+\lambda_2\mathcal{L}_\mathrm{align}+\lambda_3\mathcal{L}_\mathrm{recon}
\end{equation}
where $\lambda_1$, $\lambda_2$ and $\lambda_3$ are hyper-parameters to balance four parts of the loss function.

\paragraph{\textbf{Inference}}
In the inference process, the most related segment $\boldsymbol{\hat{\tau}}=(\hat{\tau}_\mathrm{s},\hat{\tau}_\mathrm{t})$ can be easily retrieved from the score map by:
\begin{equation}
\boldsymbol{\hat{\tau}}=(\hat{\tau_\mathrm{s}},\hat{\tau_\mathrm{t}})=\mathop{\arg\max}_\mathbf{p}M_\mathbf{p},
\end{equation}
where $M_\mathbf{p}$ is the score of the segment $\mathbf{p}$ illustrated in Equation~\ref{eq:mp}.


\begin{table*}[htb]
  \centering
  \caption{Performance comparison with UDA baselines on ActivityNet Captions. (source domain: \textit{Sport})}
  \label{tab:anet_source_sport}

  \scalebox{0.9}{
  \begin{tabular}{c|cc|cc|cc|cc|cc}
    \toprule
    \multirow{2}{*}{Method} & \multicolumn{2}{c|}{$\rightarrow$ Household} & \multicolumn{2}{c|}{$\rightarrow$ Social} & \multicolumn{2}{c|}{$\rightarrow$ Eat\&drink} & \multicolumn{2}{c|}{$\rightarrow$ Personal care} & \multicolumn{2}{c}{Average} \\
    & IoU=0.5 & IoU=0.7
    & IoU=0.5 & IoU=0.7
    & IoU=0.5 & IoU=0.7
    & IoU=0.5 & IoU=0.7
    & IoU=0.5 & IoU=0.7
    \\
    \midrule
    
    \rowcolor{mygray}Supervised-target & 40.04 & 22.77 & 44.09 & 27.43 & 36.71 & 20.09 & 38.11 & 20.31 & 39.74 & 22.65 \\

    & & & & & & & & & & \\[-2ex]
    \hline 
    & & & & & & & & & & \\[-2ex]

    Source-only & 29.21 & 14.20 & 38.04 & 20.66 & 27.89 & 13.66 & 30.40 & 14.84 & 31.39 & 15.84 \\
    MK-MMD      & 30.29 & 15.97 & 38.18 & 22.75 & 28.36 & 14.14 & 30.52 & 14.91 & 31.84 & 16.94 \\
    DeepCORAL      & 31.72 & 17.75 & 38.50 & 23.05 & 28.76 & 14.50 & 30.16 & 14.55 & 32.29 & 17.46 \\
    AFN   & 32.76 & 16.71 & 39.08 & 22.52 & 28.93 & 15.35 & 33.16 & 16.57 & 33.48 & 17.79 \\
    ACAN         & 34.27 & 16.83 & 40.27 & 22.85 & 30.76 & 15.61 & 33.95 & 16.61 & 34.81 & 17.98 \\
    \hline 
    & & & & & & & & & & \\[-2ex]
    AMDA ($\mathcal{L}_\mathrm{adv}$ only) & 34.63 & 18.14 & 40.74 & 23.27 & 31.45 & 15.96 & 33.78 & 17.02 & 35.15 & 18.60 \\
    AMDA & \textbf{38.31} & \textbf{19.78} & \textbf{42.72} & \textbf{25.49} & \textbf{32.99} & \textbf{16.83} & \textbf{36.05} & \textbf{19.65} & \textbf{37.52} & \textbf{20.44} \\

    \bottomrule
  \end{tabular}}
\end{table*}

\begin{table*}[htb]
  \centering
  \caption{Performance comparison with UDA baselines on YouCook2. (source domain: \textit{Europe})}
  \label{tab:youcook2_source_europe}

  \scalebox{0.9}{
  \begin{tabular}{c|cc|cc|cc|cc}
    \toprule
    \multirow{2}{*}{Method} & \multicolumn{2}{c|}{$\rightarrow$ America} & \multicolumn{2}{c|}{$\rightarrow$ East Asia} & \multicolumn{2}{c|}{$\rightarrow$ South Asia} & \multicolumn{2}{c}{Average} \\
    & IoU=0.3 & IoU=0.5
    & IoU=0.3 & IoU=0.5
    & IoU=0.3 & IoU=0.5
    & IoU=0.3 & IoU=0.5
    \\
    \midrule

    \rowcolor{mygray} Supervised-target & 31.52 & 17.95 & 29.12 & 17.53 & 23.31 & 11.98 & 27.98 & 15.82 \\

    & & & & & & & & \\[-2ex]
    \hline 
    & & & & & & & & \\[-2ex]

    Source-only     & 22.83 & 9.77 & 20.14 & 9.33 & 14.45 & 6.25 & 19.14 & 8.45 \\
    MK-MMD          & 23.31 & 11.28 & 21.72 & 10.05 & 15.07 & 7.13 & 20.03 & 9.49 \\
    DeepCORAL          & 24.41 & 11.91 & 22.09 & 10.55 & 15.62 & 7.55 & 20.71 & 10.00 \\
    AFN       & 24.88 & 12.11 & 23.00 & 12.76 & 14.97 & 8.07 & 20.95 & 10.98 \\
    ACAN             & 22.93 & 12.48 & 21.48 & 12.33 & 14.97 & 7.55 & 19.79 & 10.79 \\
    \hline 
    & & & & & & & & \\[-2ex]
    AMDA ($\mathcal{L}_\mathrm{adv}$ only) & 23.61 & 12.01 & 21.74 & 11.41 & 15.62 & 7.94 & 20.32 & 10.45 \\
    AMDA & \textbf{25.86} & \textbf{13.48} & \textbf{24.57} & \textbf{12.54} & \textbf{16.93} & \textbf{8.29} & \textbf{22.45} & \textbf{11.44} \\

    \bottomrule
  \end{tabular}}
\end{table*}

\section{Experiments}

\subsection{Datasets}
\paragraph{\textbf{Charades-STA}} The Charades-STA dataset \cite{gao2017tall} is an extension of the Charades dataset \cite{sigurdsson2016hollywood}, comprising 10k videos of indoor activities with an average duration of 30 seconds. It encompasses 15 indoor scenes, out of which we select the five primary scenes (accounting for approximately 60\% of the data) for evaluation, namely, \textit{Kitchen}, \textit{Bedroom}, \textit{Living room}, \textit{Bathroom}, and \textit{Entryway}. We leverage the largest scene, \textit{Kitchen}, as the source domain and aim to transfer knowledge to the remaining four scenes.

\paragraph{\textbf{ActivityNet Captions}} 
The ActivityNet Captions dataset \cite{krishna2017dense} is built upon the ActivityNet dataset \cite{caba2015activitynet} and includes 20k untrimmed videos from YouTube with 100k caption annotations. This dataset categorizes over 200 types of human activities into five scenes: \textit{Sport}, \textit{Social}, \textit{Eat\&drink}, \textit{Household}, and \textit{Personal care}. We select the biggest scene \textit{Sport} (which has the most number of annotations) as the source domain and one of the other four scenes as the target domain, respectively.

\paragraph{\textbf{YouCook2}} The YouCook2 dataset \cite{zhou2018towards} is one of the largest task-oriented instructional video datasets in the vision community. It contains 2000 long untrimmed videos from 89 cooking recipes. These recipes are divided by region: \textit{Europe}, \textit{America}, \textit{East Asia}, and \textit{South Asia}. We choose the biggest scene \textit{Europe} as the source domain and each of the other three scenes as the target domain, respectively.

\begin{table}[htbp]
  \centering
  \caption{Ablation study of AMDA model. (\textit{Sport} $\rightarrow$ \textit{Household} in ActivityNet Captions)}
  \label{tab:ablation_anet}
  \scalebox{0.95}{
  \begin{tabular}{@{}c|ccc|cc@{}}
  \toprule
  Row & $\mathcal{L}_\mathrm{adv}$ & $\mathcal{L}_\mathrm{align}$ & $\mathcal{L}_\mathrm{recon}$ & IoU=0.5 & IoU=0.7 \\ 
  \midrule
  1 & - & - & - & 29.21 & 14.20 \\
  2 & - & - & \checkmark & 31.95 & 17.63 \\
  3 & - & \checkmark & - & 33.35 & 16.66 \\
  4 & \checkmark & - & - & 34.63 & 18.14 \\
  5 & - & \checkmark & \checkmark & 35.22 & 18.40 \\
  6 & \checkmark & - & \checkmark & 36.16 & 19.06 \\
  7 & \checkmark & \checkmark & - & 36.04 & 19.17 \\
  8 & \checkmark & \checkmark & \checkmark & \textbf{38.31} & \textbf{19.78} \\ \bottomrule
  \end{tabular}}
  \end{table}

\subsection{Implementation Details}
We extract visual features by the pre-trained C3D model\cite{tran2015learning} and extract textual features by the GloVe vectors\cite{pennington2014glove}, which is consistent with previous methods\cite{zhao2021cascaded, zhang2020span}. The video length is uniformly sampled to 32. The dimension of all hidden layers is 256. In the training phase, we employ AdamW\cite{loshchilov2017decoupled} optimizer with weight decay of 1e-6. The learning rate is set to 1e-3 with cosine annealing strategy\cite{loshchilov2016sgdr}. The model is trained for 50 epochs with a batch size of 64. The dropout is set as 0.4 to prevent overfitting. The hyper-parameters $\lambda_1$, $\lambda_2$, $\lambda_3$ are set to 0.5, 0.2, 0.5, respectively. The margin $\Delta$ is 0.3, and the random ratio $\beta$ of the video mask is 0.2.

\subsection{Evaluation Metrics}
Following previous work, we adopt "R@$n$, IoU=$m$" as our evaluation metrics. This metric represents the percentage of sentence queries, at least one of whose top-n retrieved segments has an IoU (Intersection over Union) larger than m with the ground truth. In our experiment, we use $n=1$ and $m\in\{0.3, 0.5, 0.7\}$. 

\subsection{Performance Evaluation}
\label{sec:performance}
Under the UDA setting, we evaluate the performance of the proposed AMDA model in transferring knowledge from the labeled source domain to the unlabeled target domain. We compare AMDA with \textbf{Source-only}, \textbf{Supervised-target} and several traditional UDA methods (\textbf{MK-MMD} \cite{long2015learning}, \textbf{DeepCORAL} \cite{sun2016deep}, \textbf{AFN} \cite{xu2019larger}), and \textbf{ACAN} \cite{xu2022aligning} to show the effectiveness of the AMDA model. We use Base Model to train the \textbf{Source-only} (\textbf{Supervised-target}) setting with only source domain (both source and target domain) supervised data. All the traditional UDA methods are employed for both single-modal and multi-modal features, which is consistent with our adversarial loss $\mathcal{L}_\mathrm{adv}$.

The overall evaluation results on Charades-STA, ActivityNet Captions and YouCook2 are presented in Table~\ref{tab:charades_source_kitchen}, \ref{tab:anet_source_sport} and \ref{tab:youcook2_source_europe}, respectively. The results of \textbf{Supervised-target} are shown in the gray background, denoting the upper bound performance of the UDA setting. The best UDA results are given in \textbf{bold}. These experimental results reveal some notable points listed as follows:

\begin{compactitem}
\item In all three datasets, an obvious performance gap exists between \textbf{Source-only} and \textbf{Supervised-target} methods, which explicitly shows the domain gap between the source and target scene. This result proves the necessity of the cross-scene domain adaptation task.
\item Our proposed AMDA method outperforms all the traditional UDA methods on all three datasets. Especially on Charades-STA and ActivityNet Captions, the performance of AMDA is close to the supervised upper bound. Compared to the traditional UDA methods, the AMDA takes both modalities into account to learn semantic-aligned representations and pays more attention to the temporal information, which leads to success in this task.
\item As expected, we observe that the AMDA achieves a relatively lower performance gain on the YouCook2 dataset. This is because YouCook2 collects long videos of complex cooking steps, whose query segments are relatively shorter, contributing to the hardness of the cross-scene domain adaptation. Nonetheless, our model still shows effectiveness and beats other traditional UDA methods.
\end{compactitem}

\begin{figure}[tb]
\includegraphics[width=\linewidth]{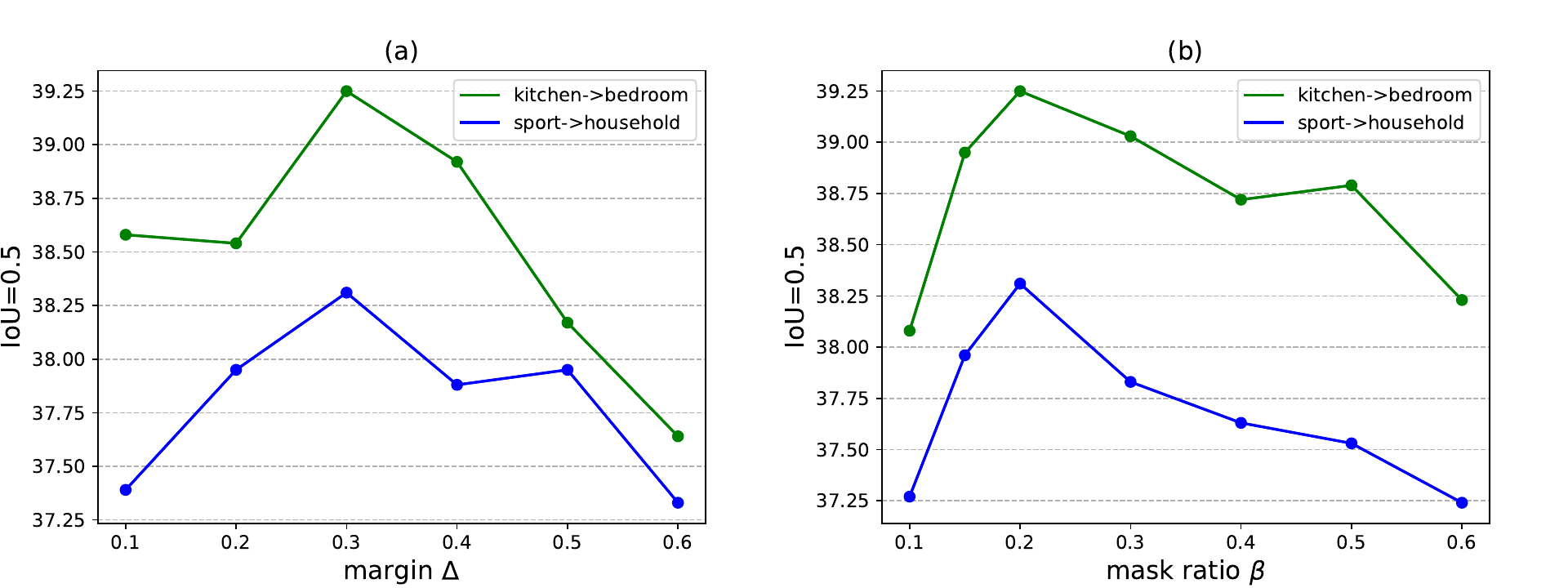}
  \caption{Analysis on margin $\Delta$ and mask ratio $\beta$.}
  \label{fig:margin_ratio}
\end{figure}

\subsection{Ablation Study\footnote[1]{Qualitative analysis and other additional experiments can be found in the appendix.}}

\paragraph{\textbf{Effect of Three Modules in AMDA}}
In order to explore the effectiveness of the three modules in AMDA, i.e., adversarial domain discriminator ($\mathcal{L}_\mathrm{adv}$), cross-modal feature alignment ($\mathcal{L}_\mathrm{align}$), and masked video reconstruction ($\mathcal{L}_\mathrm{recon}$), we conduct the ablation study on ActivityNet Captions, and the source/target domain is \textit{sport}/\textit{household}. When each of the three modules is used separately (in Row 2, 3, 4), $\mathcal{L}_\mathrm{adv}$ achieves a larger improvement than the other two modules, which proves the importance and effectiveness of $\mathcal{L}_\mathrm{adv}$ in reducing the domain gap between different domains. Meanwhile, the results (in Rows 6, 7, 8) show that $\mathcal{L}_\mathrm{align}$ and $\mathcal{L}_\mathrm{recon}$ are also indispensable parts of the AMDA method. 



\paragraph{\textbf{Analysis on Margin $\Delta$ and Mask Ratio $\beta$}}
The margin $\Delta$ in the cross-modal feature alignment module denotes how further the negative video-query pairs will be pushed away in the semantic space. The distance between negative videos and queries should not be too large in the UDA setting since there could exist domain-related semantics between them. Figure~\ref{fig:margin_ratio}(a) shows that our model performs better when $\Delta=0.3$.

In the masked video reconstruction module, the original video is randomly masked at a ratio of $\beta$. The module cannot learn enough temporal semantic relations if $\beta$ is too small. On the other hand, $\beta$ also cannot be too large to save enough useful information for the masked video. In Figure~\ref{fig:margin_ratio} (b), the resulting curve indicates that $\beta=0.2$ should be a good choice.

\paragraph{\textbf{Analysis on Adversarial Domain Discriminator}}
The adversarial domain discriminator has been employed in single-modal tasks \cite{ganin2016domain, tzeng2017adversarial}. It is intriguing to investigate whether this technique can be extended to multi-modal tasks. We conducted an evaluation of the discriminator ($\mathcal{L}_\mathrm{adv}$) on visual, textual, and multi-modal features, as illustrated in Table~\ref{tab:ablation_adv}. The results demonstrate that the domain discriminator performs well when applied to each modality individually, and it achieves even better performance when applied to all modalities jointly.

\begin{table}[htbp]
    \centering
    \caption{Ablation study of Adversarial Domain Discriminator. (\textit{Sport} $\rightarrow$ \textit{Household} in ActivityNet Captions)}
    \label{tab:ablation_adv}
    \scalebox{0.9}{
    \begin{tabular}{c|ccc|cc}
    \toprule
    Row &
        \begin{tabular}[c]{@{}c@{}}Visual \\ Feature\end{tabular} &
        \begin{tabular}[c]{@{}c@{}}Textual \\ Feature\end{tabular} &
        \begin{tabular}[c]{@{}c@{}}Multi-modal \\ Feature\end{tabular} &
        IoU=0.5 &
        IoU=0.7 \\ \midrule
    1 & - & - & - & 29.21 & 14.20 \\
    2 & - & \checkmark & - & 32.78 & 17.41 \\
    3 & \checkmark & - & - & 31.53 & 16.57 \\
    4 & \checkmark & \checkmark & - & 33.56 & 17.55 \\
    5 & - & - & \checkmark & 33.57 & 17.05 \\
    6 & \checkmark & \checkmark & \checkmark & 34.63                & 18.14                \\ \bottomrule
    \end{tabular}}
\end{table}

\section{Conclusion and Limitation}
This paper investigates the UDA setting across video scenes for the TVG task, where only the source domain has labeled temporal boundaries, but the target domain does not. To address this issue, we propose a novel AMDA method to enhance performance on new target scenes. Extensive experiments on three datasets demonstrate the effectiveness of our proposed method. 

In this work, we test our proposed AMDA method only based on a simple two-stage grounding architecture. Given the architecture similarity of all the SOTA two-stage models, we may be able to extrapolate the results to other models, but further work is needed to confirm the effectiveness of AMDA on other model architectures.

We explore a potential approach to adapting the model's knowledge to new scenes in real-world application scenarios. However, this is not a definitive solution to all the challenges that may arise in such applications. Further research is necessary to facilitate the practical implementation of temporal video grounding tasks. We will make our code available to the community for replication and extension.

\bibliography{main.bbl}

\appendix

\begin{table*}[htb]
    \centering
    \caption{Comparison between Source-only and Target-only on Charades-STA. (source domain: \textit{Kitchen})}
    \label{tab:charades_source_target}
    
    \scalebox{0.9}{
    \begin{tabular}{c|cc|cc|cc|cc|cc}
        \toprule
        \multirow{2}{*}{Method} & \multicolumn{2}{c|}{-> Bedroom} & \multicolumn{2}{c|}{-> Living room} & \multicolumn{2}{c|}{-> Bathroom} & \multicolumn{2}{c|}{-> Entryway} & \multicolumn{2}{c}{Average}\\
        & IoU=0.5 & IoU=0.7
        & IoU=0.5 & IoU=0.7
        & IoU=0.5 & IoU=0.7
        & IoU=0.5 & IoU=0.7
        & IoU=0.5 & IoU=0.7
        \\
        \midrule
        
        Source-only     & 28.82 & 13.31 & 36.85 & 20.96 & 35.58 & 17.01 & 21.15 & 11.22 & 30.60 & 15.63 \\
        Target-only          & \textbf{36.74} & \textbf{17.65} & \textbf{38.43} & \textbf{21.25} & \textbf{38.65} & \textbf{20.68} & \textbf{29.04} & \textbf{14.20} & \textbf{35.72} & \textbf{18.45} \\

        \bottomrule
    \end{tabular}}
    \end{table*}

\section{Analysis on Domain Gap}
We present the analysis results of the domain gap in the Charades-STA dataset, which demonstrate the rationality and necessity of our work.

\begin{figure*}[htb]
    \includegraphics[width=\linewidth]{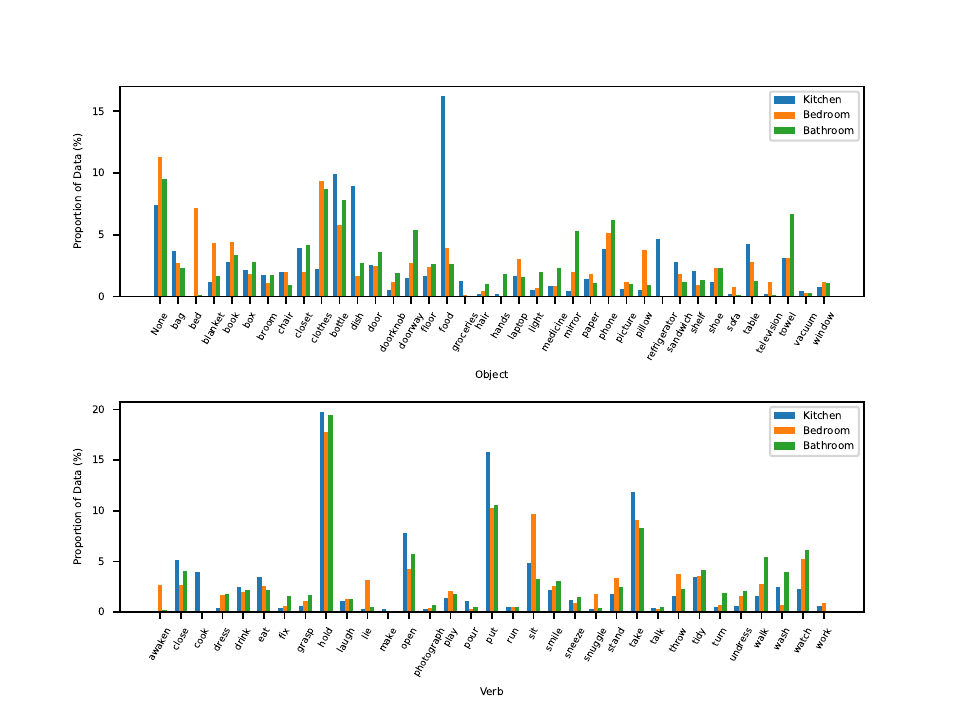}
        \caption{Class distribution of "object" and "verb", respectively, for three main scenes (Kitchen, Bedroom, and Bathroom) in Charades-STA dataset.}
        \label{fig:charades_distribution}
    \end{figure*}

\paragraph{\textbf{Class Distribution}}
The Charades-STA dataset \cite{gao2017tall} comprises 10,000 videos from 15 indoor scenes. In our work, we focus on the largest five scenes for our experiments, with the largest scene---\textit{Kitchen}---serving as the source domain. To measure the impact of scene factors in different scenes, we examine the class distribution of "object" and "verb" for three scenes in particular: \textit{Kitchen}, \textit{Bedroom}, and \textit{Bathroom}. As shown in Figure~\ref{fig:charades_distribution}, while all three scenes contain a high number of objects and verbs, the frequency of specific objects and verbs differs significantly across scenes. For instance, the objects "dish" and "food" are most commonly found in the \textit{Kitchen} scene, while the objects "bed" and "pillow" are most commonly found in the \textit{Bedroom} scene. Such differences in class distribution exacerbate the domain shift between scenes and impose strong scene-related factors on the learned knowledge of supervised models.

\paragraph{\textbf{Performance Drop}}
Given that the superior performance of the \textbf{Supervised-target} setting over the \textbf{Source-only} setting may be attributed to the larger amount of training data, we evaluate the \textbf{Target-only} setting, which only trains on the target domain, to more explicitly demonstrate the domain gap. As we use the largest scene as the source domain, the target domain has fewer training samples but achieves better performance, as shown in Table~\ref{tab:charades_source_target}, which directly comes from the domain gap. The number of samples for the five scenes (Kitchen, Bedroom, Living room, Bathroom, and Entryway) are 2663, 2581, 2355, 1157, and 952, respectively.

\section{Dataset Analysis}
\label{sec:appendix_dataset}
We conduct a survey on the mainstream datasets in TVG task, including Charades-STA \cite{gao2017tall}, ActivityNet Captions \cite{krishna2017dense}, YouCook2 \cite{zhou2018towards}, and TACoS \cite{regneri2013grounding}.

\begin{table}[htb]
    \centering
    \caption{The statistics of Charades-STA dataset.}
    \label{tab:charades_dataset}
    \scalebox{0.9}{
    \begin{tabular}{c|c|c|c}
        \toprule
        scene & \# of videos & \begin{tabular}[c]{@{}c@{}}\# of\\ annotations\end{tabular} & \begin{tabular}[c]{@{}c@{}}\# of\\ action class\end{tabular} \\
        \midrule
        \textbf{Kitchen}     & \textbf{1541} & \textbf{2663} & \textbf{156} \\
        \textbf{Bedroom}     & \textbf{1503} & \textbf{2581} & \textbf{153} \\
        \textbf{Living room} & \textbf{1460} & \textbf{2355} & \textbf{157} \\
        \textbf{Bathroom}    & \textbf{672}  & \textbf{1157} & \textbf{149} \\
        \textbf{Entryway}    & \textbf{544}  & \textbf{952}  & \textbf{153} \\
        Home Office     & 562  & 881  & 152 \\
        Closet          & 430  & 831  & 142 \\
        Dining room     & 511  & 819  & 157 \\
        Laundry room    & 435  & 681  & 150 \\
        Hallway         & 477  & 675  & 141 \\
        Stairs          & 503  & 576  & 141 \\
        Recreation room & 332  & 522  & 149 \\
        Pantry          & 291  & 511  & 145 \\
        Garage          & 290  & 493  & 148 \\
        Basement        & 192  & 286  & 148 \\
        \bottomrule
        \end{tabular}
}
\end{table}

\begin{table}[htb]
    \centering
    \caption{The statistics of ActivityNet Captions dataset.}
    \begin{tabular}{c|c|c}
    \toprule
    scene         & train samples & test samples \\
    \midrule
    sport         & 17057         & 8095         \\
    household     & 7471          & 3521         \\
    social        & 7036          & 3119         \\
    personal care & 3150          & 1515         \\
    eat \& drink  & 2707          & 1255         \\
    \bottomrule
    \end{tabular}
\end{table}

\begin{figure*}[htb]
\includegraphics[width=\linewidth]{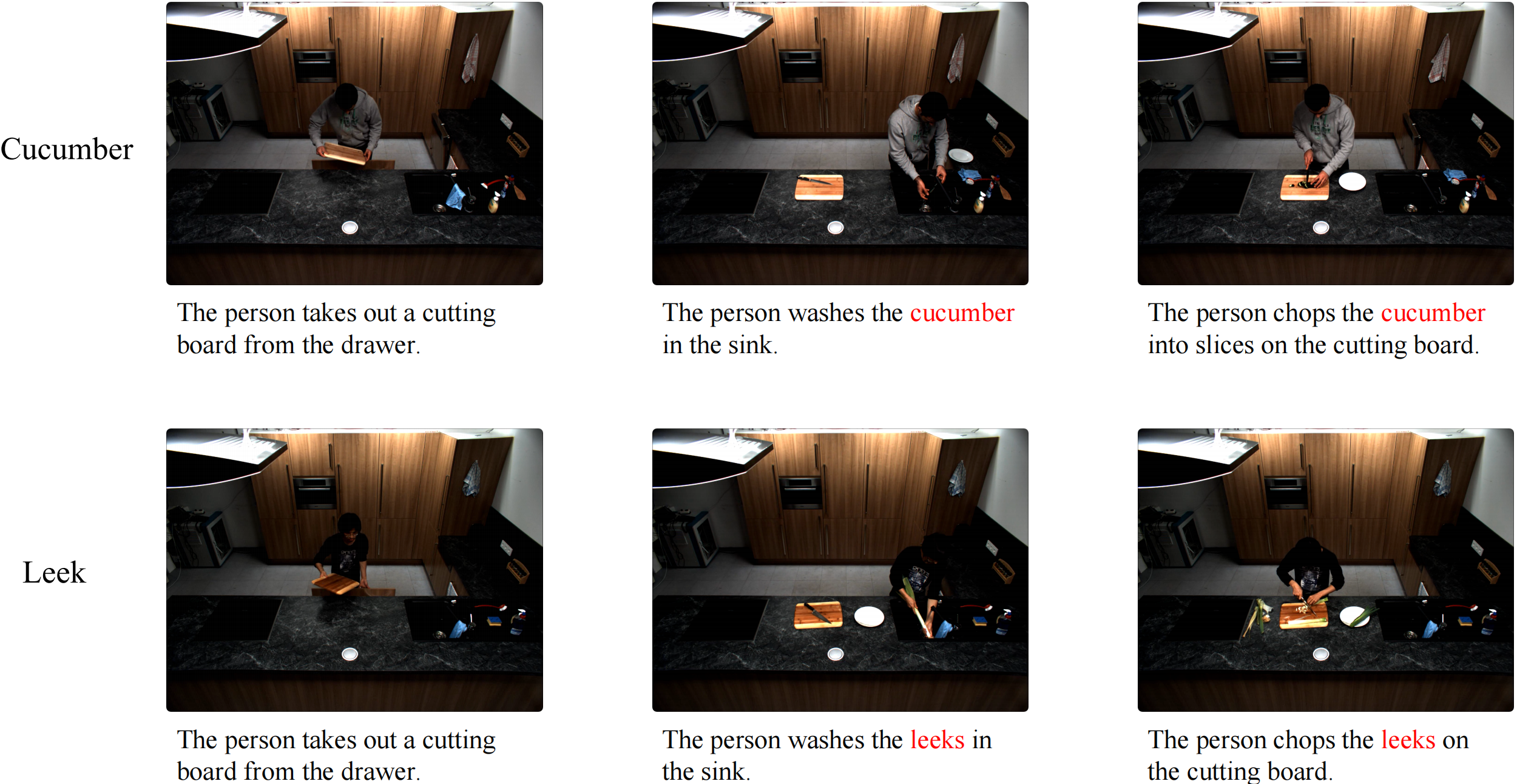}
    \caption{The whole TACoS dataset are collected in the same kitchen scene. The only difference of these videos are the food ingredients they use.}
    \label{fig:tacos}
\end{figure*}

\paragraph{Charades-STA}
The Charades-STA dataset \cite{gao2017tall} collects 10k videos from 15 indoor scenes. All the annotations can be categorized into 157 action classes. And each indoor scene will cover almost all the action classes. The statistics are given in Table~\ref{tab:charades_dataset}. In our work, we use the largest five scenes for the experiment and the largest one---\textit{Kitchen} for the source scene. In order to measure the importance of scene factors in different scenes, we take a closer look at the class distribution of "object" and "verb", respectively, for three scenes--\textit{Kitchen}, \textit{Bedroom}, and \textit{Bathroom}. As shown in Figure~\ref{fig:charades_distribution}, although all three scenes contain most objects and scenes, the frequency differs a lot across scenes. For example, the objects "dish" and "food" appear mostly in \textit{Kitchen} scene, while the objects "bed" and "pillow" are mostly in \textit{Bedroom} scene. This kind of class distribution difference aggravates the domain shift between scenes and imposes strong scene-related factors on the learned knowledge of supervised models.

\paragraph{ActivityNet Captions}
The ActivityNet Captions dataset \cite{krishna2017dense} includes 20k untrimmed videos from YouTube. This dataset categorizes over 200 types of human activities into five scenes: \textit{sport}, \textit{social}, \textit{eat\&drink}, \textit{household}, and \textit{personal care}. Here, we expand the meaning of the scene to the coarse-grained category, which also satisfies the basic logic of our UDA setting. The domain gap can be explicitly observed from the experiment result.

\paragraph{YouCook2}
The YouCook2 dataset \cite{zhou2018towards} contains 2000 long untrimmed videos from 89 cooking recipes. Since the cooking videos are all collected from kitchens, we explore the fine-grained division of scenes, which are divided by region: \textit{Europe}, \textit{America}, \textit{East Asia}, and \textit{South Asia}. The domain gap is also obvious in this scene division.

\paragraph{TACoS}
The TACoS dataset \cite{regneri2013grounding} contains 127 videos collected from the MPII Cooking Composite Activities video corpus \cite{rohrbach2012script}. Considering the wide usage of this dataset in the TVG task, we also plan to conduct experiments on it. Unfortunately, we find that all the videos are collected in the same kitchen, meaning there is no difference in scene factors in different videos. Moreover, some annotations are almost the same in different videos, as shown in Figure~\ref{fig:tacos}. Therefore, we can't obtain a reasonable division of these videos to evaluate our method.

\section{Ablation Study}

\begin{table}[htbp]
\centering
\caption{Ablation study of Adversarial Domain Discriminator. (\textit{Sport} $\rightarrow$ \textit{Household} in ActivityNet Captions)}
\label{tab:ablation_adv}
\scalebox{0.9}{
\begin{tabular}{c|ccc|cc}
\toprule
Row &
    \begin{tabular}[c]{@{}c@{}}Visual \\ Feature\end{tabular} &
    \begin{tabular}[c]{@{}c@{}}Textual \\ Feature\end{tabular} &
    \begin{tabular}[c]{@{}c@{}}Multi-modal \\ Feature\end{tabular} &
    IoU=0.5 &
    IoU=0.7 \\ \midrule
1 & - & - & - & 29.21 & 14.20 \\
2 & - & \checkmark & - & 32.78 & 17.41 \\
3 & \checkmark & - & - & 31.53 & 16.57 \\
4 & \checkmark & \checkmark & - & 33.56 & 17.55 \\
5 & - & - & \checkmark & 33.57 & 17.05 \\
6 & \checkmark & \checkmark & \checkmark & 34.63                & 18.14                \\ \bottomrule
\end{tabular}}
\end{table}

\begin{table}[htbp]
\centering
\caption{Hyperparameter Study. (\textit{Sport} $\rightarrow$ \textit{Household} in ActivityNet Captions)}
\label{tab:hyperparameter}
\scalebox{0.9}{
\begin{tabular}{c|ccc|cc}
    \toprule
    Row & $\lambda_1$ & $\lambda_2$ & $\lambda_3$ & IoU=0.5 & IoU=0.7 \\ \midrule
    1   & 0.1      & 0.2      & 0.5      & 38.14   & 19.42   \\
    2   & 0.2      & 0.2      & 0.5      & 38.42   & 19.65   \\
    3   & 0.3      & 0.2      & 0.5      & 38.08   & 19.15   \\ \hline
    4   & 0.5      & 0.1      & 0.5      & 38.28   & 19.53   \\
    5   & 0.5      & 0.2      & 0.5      & 38.31   & 19.78   \\
    6   & 0.5      & 0.3      & 0.5      & 38.10    & 19.08   \\ \hline
    7   & 0.5      & 0.2      & 0.1      & 38.22   & 19.23   \\
    8   & 0.5      & 0.2      & 0.2      & 38.53   & 19.89   \\
    9   & 0.5      & 0.2      & 0.3      & 38.09   & 19.21   \\ \bottomrule
    \end{tabular}}
\end{table}

\paragraph{\textbf{Analysis on Adversarial Domain Discriminator}}
The adversarial domain discriminator has been employed in single-modal tasks \cite{ganin2016domain, tzeng2017adversarial}. It is intriguing to investigate whether this technique can be extended to multi-modal tasks. We conducted an evaluation of the discriminator ($\mathcal{L}_\mathrm{adv}$) on visual, textual, and multi-modal features, as illustrated in Table~\ref{tab:ablation_adv}. The results demonstrate that the domain discriminator performs well when applied to each modality individually, and it achieves even better performance when applied to all modalities jointly.

\paragraph{\textbf{Hyperparameter Study}}
It is intriguing to investigate the selection of loss hyperparameters for multi-task learning. In our study, instead of performing an exhaustive hyperparameter search, we simply set different loss terms to the same level of magnitude, which is a common practice for setting loss hyperparameters. To evaluate the sensitivity of our model to the hyperparameters, we conduct training for each of $\lambda_1$, $\lambda_2$, and $\lambda_3$, by setting them to 0.1, 0.2, and 0.3, respectively. As illustrated in the Table~\ref{tab:hyperparameter}, it can be observed that the effect of loss hyperparameters (within a reasonable range) on the model's performance is not significant.


\begin{figure*}[tb]
\includegraphics[width=\linewidth]{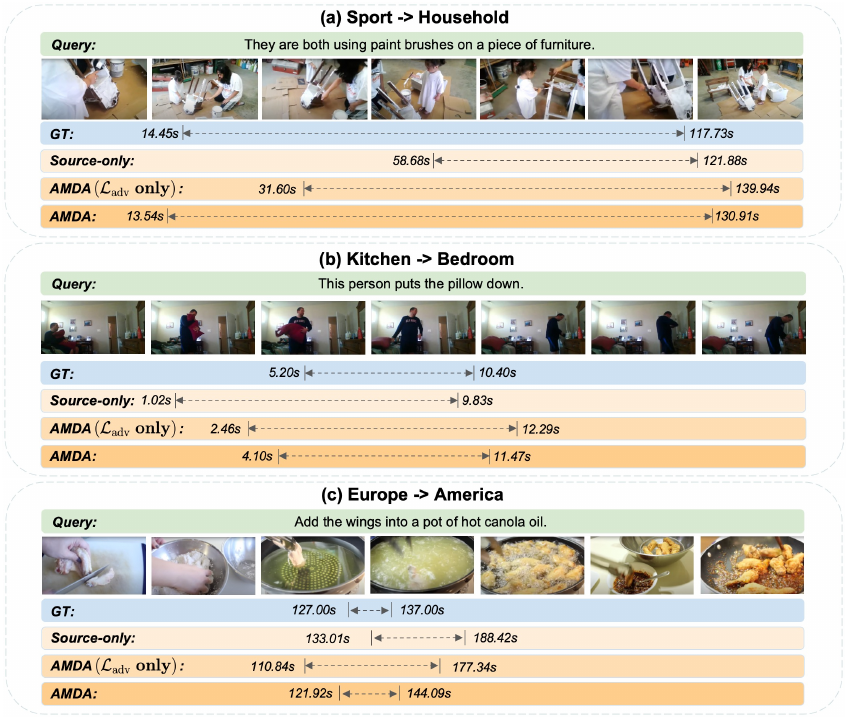}
    \caption{Qualitative results of AMDA on ActivityNet Captions (Sport$\rightarrow$Household), Charades-STA (Kitchen$\rightarrow$Bedroom), and YouCook2 (Europe$\rightarrow$America), compared with Source-only and AMDA($\mathcal{L}_\mathrm{adv}$ only) methods.}
    \label{fig:qualitative}
\end{figure*}

\begin{figure*}[tb]
\includegraphics[width=\linewidth]{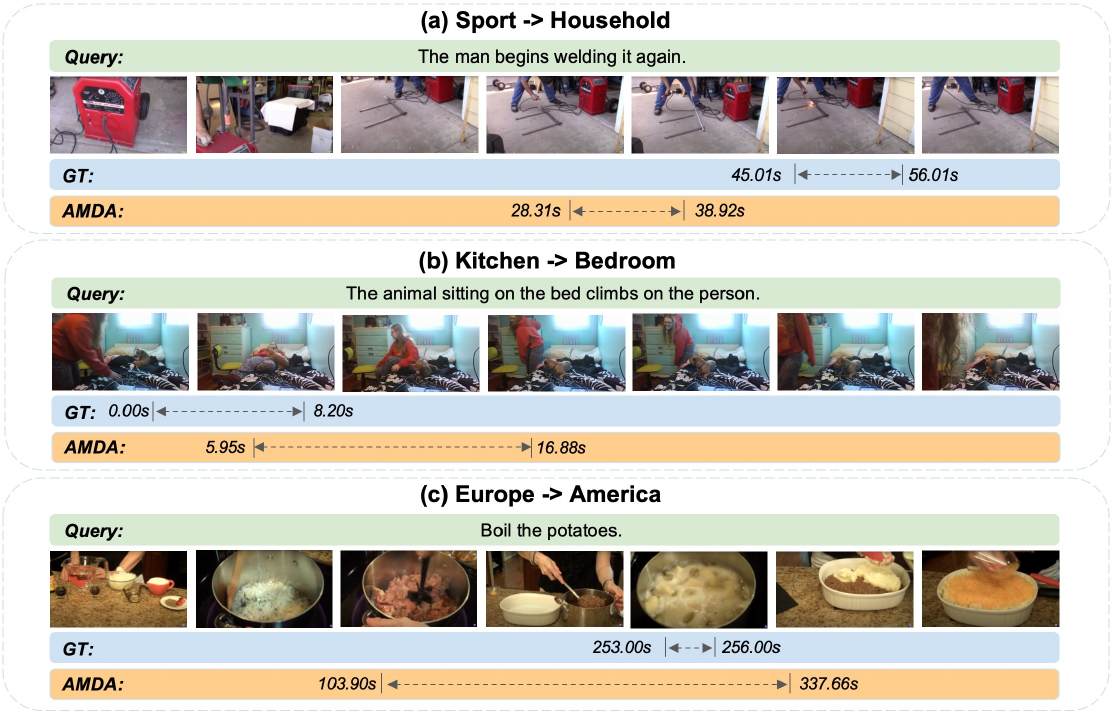}
    \caption{Failure cases of AMDA method on three datasets.}
    \label{fig:failure_case}
\end{figure*}

\section{Qualitative Analysis}
\paragraph{\textbf{Qualitative Comparison}} 
To assess the AMDA method's performance qualitatively, we provide examples from the three datasets in Figure \ref{fig:qualitative}. In these instances, the source-only method fails to accurately localize boundaries due to its inability to learn from the target domains. In contrast, the AMDA model can make more precise predictions with the aid of unlabeled target data. Furthermore, these examples demonstrate that the AMDA model can predict both long-range and short-range segments effectively.

\paragraph{\textbf{Failure Cases}}
Despite its outstanding performance compared to other UDA methods, AMDA may still fail in certain cases, as illustrated in Figure~\ref{fig:failure_case}. (a) In this instance, the query is asking for the second occurrence of action "weld the item," while AMDA returns the answer segment for the first occurrence of welding. This indicates that our model has yet to fully explore temporal semantics. (b) In this case, an out-of-domain subject (an animal) appears in the query, which hinders the prediction. (c) In this instance, the target segment is too short to be accurately detected, and the model appears to generate predictions randomly.

\end{document}